\documentclass{bmvc2k}

\title{Recoding Color Transfer as a Color Homography}

\addauthor{Han Gong}{http://www2.cmp.uea.ac.uk/~ybb15eau/}{1}
\addauthor{Graham D. Finlayson}{g.finlayson@uea.ac.uk}{1}
\addauthor{Robert B. Fisher}{http://homepages.inf.ed.ac.uk/rbf/}{2}

\addinstitution{
 School of Computing Sciences\\
 University of East Anglia\\
 Norwich, UK
}
\addinstitution{
 School of Informatics\\
 University of Edinburgh\\
 Edinburgh, UK
}

\runninghead{Gong, Finlayson, Fisher}{Recoding Color Transfer as a Color Homography}

\def\eg{\emph{e.g}\bmvaOneDot}

\def\ie{\emph{i.e}\bmvaOneDot}
\def\etal{\emph{et al}\bmvaOneDot}

\usepackage{amssymb}
\usepackage{caption}
\usepackage{multicol}
\usepackage{textcomp}

\usepackage{booktabs}
\usepackage{soul}

\usepackage[linesnumbered]{algorithm2e}
\newcommand{\norm}[1]{\left\lVert #1 \right\rVert}

\begin{document}

\maketitle

\begin{abstract}
Color transfer is an image editing process that adjusts the colors of a picture to match a target picture's color theme. A natural color transfer not only matches the color styles but also prevents after-transfer artifacts due to image compression, noise, and gradient smoothness change. The recently discovered color homography theorem proves that colors across a change in photometric viewing condition are related by a homography. In this paper, we propose a color-homography-based color transfer decomposition which encodes color transfer as a combination of chromaticity shift and shading adjustment. A powerful form of shading adjustment is shown to be a global shading curve by which the same shading homography can be applied elsewhere. Our experiments show that the proposed color transfer decomposition provides a very close approximation to many popular color transfer methods. The advantage of our approach is that the learned color transfer can be applied to many other images (\eg other frames in a video), instead of a frame-to-frame basis. We demonstrate two applications for color transfer enhancement and video color grading re-application. This simple model of color transfer is also important for future color transfer algorithm design.
\end{abstract}

%------------------------------------------------------------------------- 
\section{Introduction}
\label{sec:intro}
Adjusting the color style of pictures/frames is one of the most common tasks in professional photo editing as well as video post-production. Artists would often choose a desired target picture and manipulate the other pictures to match their target color style. This process is called color transfer. An example of color transfer between a source image and a target image is shown in Figure~\ref{fig:demo}. Typically, this color tuning process requires artists to delicately adjust for multiple properties such as exposure, brightness, white-point, and color mapping. These adjustments are also interdependent, \ie when aligning an individual property this may cause the others to become misaligned. For rendered images, some artifacts (\eg JPEG block edges) may appear after color adjustment. It is therefore desirable to automate this time-consuming task.

Example-based color transfer was first introduced by Reinhard \etal~\cite{ReinhardTransfer}. Since then, much further research~\cite{Pitie1,Pitie2,Pouli,Nguyen} has been carried out. A recent discovery of the color homography theorem reveals that colors across a change in viewing condition (illuminant, shading or camera) are related by a homography~\cite{PICS2016,CIC2016}. In this paper, we propose a general model, based on the color homography theorem, to approximate different color transfer results. In our model, we decompose any color transfer into a chromaticity mapping component and a shading adjustment component. We also show that the shading adjustment can be reformulated by a global shading curve through which the shading homography can be applied elsewhere. Our experiments show that our model produces very close approximations to the original color transfer results. We believe that our color transfer model is useful and fundamental for developing simple and efficient color transfer algorithms. That is a trained model can be applied to every frame in a video fragment rather than needing frame-by-frame adjustment. This decomposition also enables users to amend the imperfections of a color transfer result or simply extract the desired effect.

Our paper is organized as follows. We review the popular color transfer methods and the color homography theorem in \S\ref{sec:background}. Our color transfer decomposition is described in \S\ref{sec:decoding}. We show our evaluation and applications in \S\ref{sec:results}. Finally, we conclude in \S\ref{sec:conclusion}.

\section{Background}
\label{sec:background}
%In this section, we review the popular color transfer methods and briefly describe the color homography theorem.
\subsection{Color transfer}
Example-based color transfer was first introduced by Reinhard \etal~\cite{ReinhardTransfer}. Their method assumes that the color distribution in $l\alpha\beta$ color space is a normal distribution. They map a source image to its target so that their color distributions have the same mean and variance in $l\alpha\beta$ color space.
Pitie \etal~\cite{Pitie1} proposed an iterative color transfer method that rotates and shifts the color distribution in 3D until the distributions of the two images are aligned. The rotation matrix is random over all possible angular combinations. This method was later improved by adding a gradient preservation constraint to reduce after-transfer artifacts~\cite{Pitie2}.
Pouli and Reinhard~\cite{Pouli} adopted a progressive histogram matching in L*a*b* color space. In their color transfer method, users can specify the level of color transfer (i.e. partial color transfer) between two images. Their algorithm also addresses the difference in dynamic ranges of between two images. 
Nguyen \etal~\cite{Nguyen} proposed an illuminant-aware and gamut-based color transfer. A white-balancing step is first performed for both images to remove color casts caused by different illuminations. A luminance matching is later performed by histogram matching along the ``gray'' axis of RGB. They finally adopt a 3D convex hull mapping, which contains scale and rotation operations, to ensure that the color-transferred RGBs are still in the space of the target RGBs.
There are some other approaches (\eg \cite{an2010user,wu2013content,tai2005local,kagarlitsky2009piecewise,chang2015palette}) that solve for several local color transfers rather than a single global color transfer. In this paper, we focus on global color transfer.

Pitie \etal~\cite{MKL_ct} proposed a color transfer approximation by a 3D affine mapping. The linear transform minimizes the amount of changes in color and preserves the monotonicity of intensity changes. However, it is based on the assumption that the color distributions of the two images are both normal distributions. It also does not well approximate the shading change of a color transfer.

\subsection{Color homography}
The color homography theorem~\cite{PICS2016,CIC2016} shows that chromaticities across a change in capture conditions (light color, shading and imaging device) are a homography apart. Let us map an RGB $\underline{\rho}$ to a corresponding RGI (red-green-intensity) \underline{c} using a $3\times 3$ full-rank matrix $C$:
\begin{equation}
\begin{array}{c}
\underline{\rho}^\intercal C=\underline{c}^\intercal\\
\;\\
\left [
\begin{array}{c}
R\\
G\\
B
\end{array}
\right ]^\intercal
\left [
\begin{array}{ccc}
1 & 0 & 1\\
0 & 1 & 1\\
0 & 0 & 1
\end{array}
\right ]
=
\left [ 
\begin{array}{c}
R\\
G\\
R+G+B
\end{array}
\right ]^\intercal

\end{array}
\label{eq:chromaticity_conversion}
\end{equation}
The $r$ and $g$ chromaticity coordinates are written as ${r=R/(R+G+B)} \;,\;{g=G/(R+G+B)}$. We interpret the right-hand-side of Equation~\ref{eq:chromaticity_conversion} as a homogeneous coordinate and we have $
\underline{c}\propto \left [
\begin{array}{ccc} r&g&1
\end{array}
\right]^\intercal
$.
When the shading is fixed, it is well-known that across a change in illumination or a change in device, the corresponding RGBs are related by a $3 \times 3$ linear transform M that $\underline{\rho}^\intercal M = \underline{\rho}'^\intercal$ where $\underline{\rho}'$ is the corresponding RGBs under a second light or captured by a different camera~\cite{MARIMONT.WANDELL,MALONEY86B}. Clearly, $H=C^{-1}MC$ maps colors in RGI form between illuminants. Due to different shading, the RGI triple under a second light is written as $\underline{c}'^\intercal=\alpha\underline{c}^\intercal H$, where $\alpha$ denotes the unknown scaling. Without loss of generality let us interpret \underline{c} as a homogeneous coordinate i.e. assume its third component is 1. Then, $[r'\;g']^\intercal=H([r\;g]^\intercal)$ (rg chromaticity coordinates are a homography $H()$ apart).
%In color, homography is applied to map 3D colors in a shading independent manner.
%We can envisage solving for the homography as finding the best linear mapping for chromaticities in different viewing conditions. In another word, we map 2D chromaticities to corresponding 3D rays, apply the $3\times 3$ matrix, and then recompute chromaticities.

\section{Recoding color transfer as a color homography}
\label{sec:decoding}
Color transfer can often be interpreted as re-rendering an image with respect to real physical scene changes (e.g. from summer to autumn) and/or illumination. Recent work~\cite{MKL_ct} approximates the effect of global color transfer by a 3D affine mapping. We propose that, in general, we can better approximate most global color transfer algorithms as a color homography transfer. The color homography theorem shows that the same scene under an illuminant (or camera) change will result in two images a homography apart. We propose that a global color transfer can be decomposed into a linear chromaticity mapping and a shading adjustment. This concise form enables us to efficiently replicate the originally slow color transfer and re-apply it to many other images (\eg the frames of a video fragment). 

\begin{figure}[htb!]
    \begin{center}
        \includegraphics[width=\linewidth]{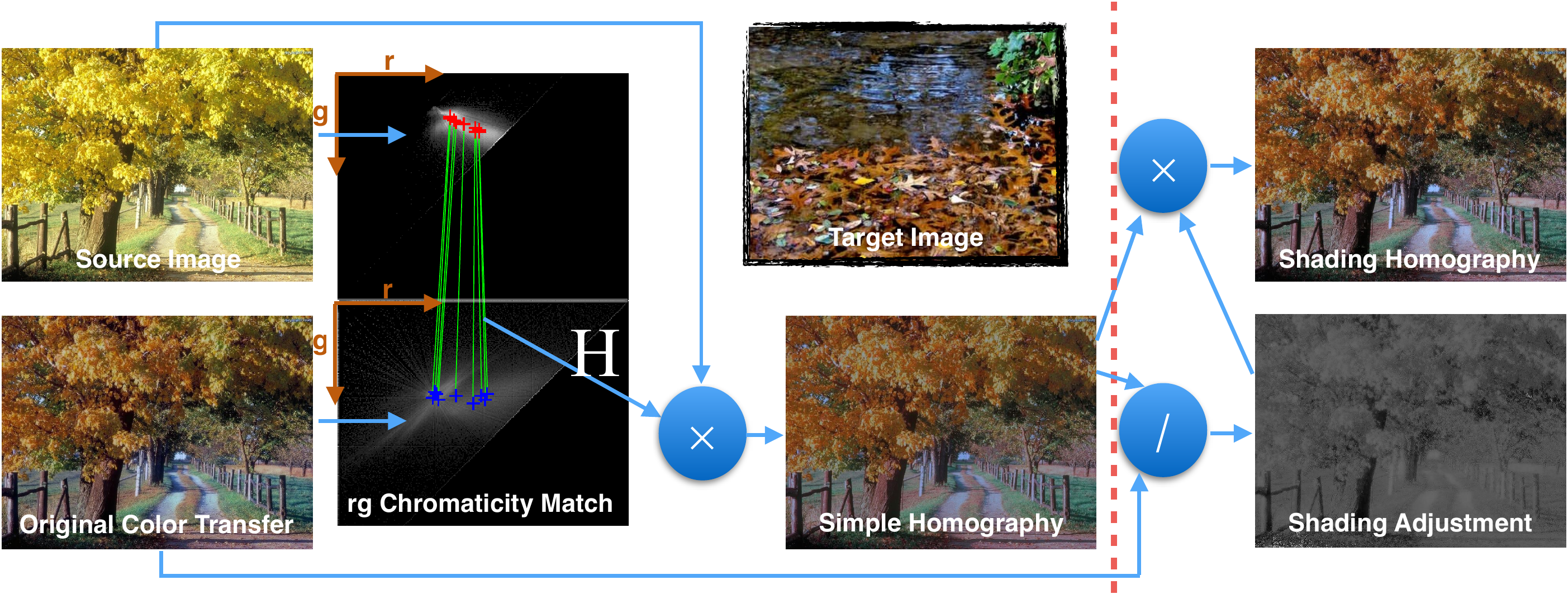}
    \end{center}
    \caption{Pipeline of color-homography-based color transfer decomposition. The red dashed line divides the pipeline into two steps: 1) {\it Simple} homography. The rg chromaticities of the source image and the original color transfer image (by \cite{Pitie2}) are matched according to their chromaticity locations (\eg the green lines), from which we estimate a color homography matrix $H$ and use $H$ to transfer the source image. 2) {\it Shading} homography. The shadings are aligned between the {\it simple} homography result and the original color transfer result by a least-squares method. The per-pixel product of the {\it simple} homography result and the shading adjustment gives a close color transfer approximation.}
    \label{fig:demo}
    \vspace{-5pt}
\end{figure}
Throughout the paper we denote the source image by $I_s$ and the original color transfer result by $I_t$. Given $I_s$ and $I_t$, the aim of color transfer decomposition is to find a general model that reproduces the color theme change from $I_s$ to $I_t$. Figure~\ref{fig:demo} shows our two-step color transfer decomposition: 1) Chromaticity mapping estimation ({\it simple} homography). The source image is chromaticity transferred by applying a color homography transform estimated from the corresponding rg chromaticities of the source image and the original color transfer result. 2) Shading adjustment estimation ({\it shading} homography). A further shading adjustment is estimated by finding a least-squares solution that aligns the shadings of the original color transfer output and the chromaticity transferred image. These procedures are explained in detail in the following.

\subsection{Color Homography Color Transfer Model}
We start with the outputs of the prior-art algorithms. Assuming we relate $I_s$ to $I_t$ with a pixel-wise correspondence, we represent the RGBs of $I_s$ and $I_t$ as two $n\times 3$ matrices $A$ and $B$ respectively where $n$ is the number of pixels. These $n\times 3$ matrices can be reconstituted into the original image grids. The chromaticity mapping is modeled as a $3\times 3$ linear transform but because of the relative positions of light and surfaces there might also be per-pixel shading perturbations. Assuming the Lambertian image formation an accurate physical model, 
\begin{equation}
DAH\approx B
\label{eq:image_model}
\end{equation}
where $D$ is an $n\times n$ diagonal matrix of shading factors and $H$ is a $3\times 3$ chromaticity mapping matrix. A color transfer can be decomposed into a diagonal shading matrix $D$ and a homography matrix $H$. The homography matrix $H$ is a global chromaticity mapping. The matrix $D$ can be seen as a change of surface reflectance or position of illuminant.

According to the color homography model, we define two color transfer decomposition models. In {\it simple} homography transfer, the output image is a homography from the input which only contains a chromaticity mapping. It preserves the shading of the source image and does not include the shading of the original color transfer result. In {\it shading} homography transfer, the output also incorporates the best shading factors which restore the shading of the original color-transfer output. By solving for the homography $H$, the {\it simple} and {\it shading} homography transfers are defined as:
\begin{equation}
B_{simple}=AH\approx B
\label{eq:simple}
\end{equation}
\begin{equation}
    B_{shading}=DAH = DB_{simple} \approx B
\label{eq:shading}
\end{equation}
An example of the two color transfer models are shown in Figure~\ref{fig:demo}. When recoding a color transfer, these two recoding models provide an additional flexibility as some users may only attempt to extract the chromaticity mapping.

\subsection{Chromaticity mapping estimation}
In color correction, Equation~\ref{eq:image_model} is solved by using Alternating Least-Squares (ALS)~\cite{PICS2016,CIC2016,ALS} illustrated in Algorithm~\ref{alg:als}.
\begin{algorithm}
\SetAlgoLined
    $i=0$, $\min_{D^0} \norm{D^0A-B}_F$, $A^0=D^0 A$\;
    \Repeat{$\norm{A^{i}-A^{i-1}}_F<\epsilon$ % \textbf{And}\;D^i = {\cal I}_{n \times n}\;\textbf{And}\;H^{i} = {\cal I}_{3 \times 3}$
    }
    {
    $i=i+1$\;
    $\min_{H^i} \norm{A^{i-1}H^i-B}_F$\;
    $\min_{D^i} \norm{D^iA^{i-1}H^i-B}_F$\;
    $A^{i}=D^iA^{i-1}H^i$\;
    }
\caption{Homography from alternating least-squares}
\label{alg:als}
\end{algorithm}

The effect of the individual $H^i$ and $D^i$ can be merged into a single matrix $D=\prod_{i} D^i$ and $H=\prod_{i} H_i$ (where the product is taken by post-multiplying matrices). $\norm{.}_{F}$ denotes the Frobenius norm and $H^i$ and $D^i$ are found using the closed form Moore-Penrose inverse.
In color transfer, we choose to minimize the rg chromaticity (\ie normalized RG) difference between two images because any non-zero RGB can be mapped to the range of $[0,1]$. To achieve this, we modify Equation~\ref{eq:image_model} as
\begin{equation}
    Dh(AC)H_{rg}\approx h(BC)
\label{eq:image_model_rg}
\end{equation}
where $C$ is the $3 \times 3$ RGB-to-RGI conversion matrix defined in Equation~\ref{eq:chromaticity_conversion}, $h$ is a function that converts each RGI intensity (matrix row) to their homogeneous coordinates (by dividing RGI by I which makes all elements of the 3$^{\text{rd}}$ column 1), $H_{rg}$ is the homography matrix that minimizes rg chromaticity difference. The homography matrix $H$ for color transfer is related with $H_{rg}$ by $H = CH_{rg}C^{-1}$. As the under-saturated pixels with zero RGBs contain no color information, they are excluded from the computation.

\begin{figure}[htb!]
\begin{center}
  \includegraphics[width=\linewidth]{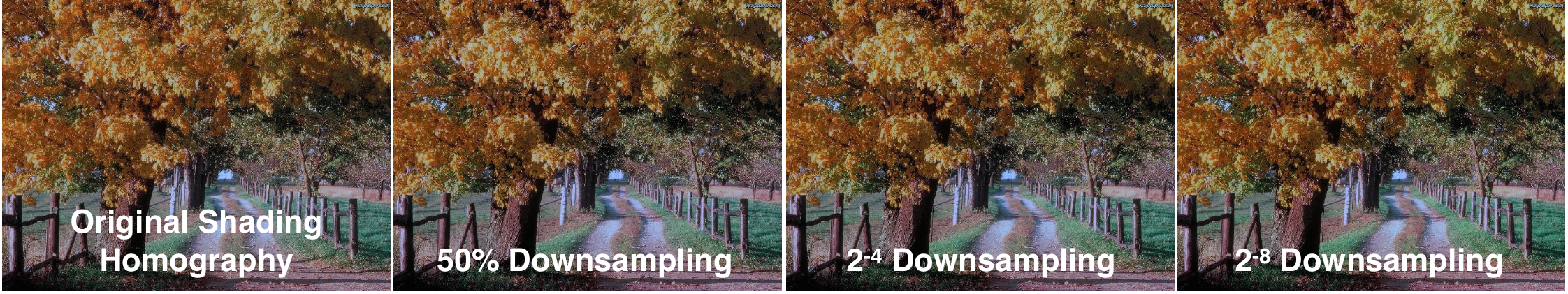}
\end{center}
\caption{Down-sampled images for chromaticity mapping estimation (shading adjustment is still estimated using full-resolution images). The sizes of $I_s$ and $I_t$ are reduced by the corresponding factors. Image down-sampling barely affects the color transfer approximation quality such that even two $3 \times 4$ thumbnails are sufficient for getting the approximation result shown in the right-most figure ($2^{-8}$ down-sampling).}
  \label{fig:downsampling}
    \vspace{-10pt}
\end{figure}

To reduce the computational cost, it is possible to estimate $H$ with down-sampled images. We find that image down-sampling barely affects our chromaticity mapping quality. An example is shown in Figure~\ref{fig:downsampling}. Depending on the content of image, the minimum down-sampling factor for estimating $H$ may vary.

\subsection{Shading adjustment estimation}
A chromaticity-transferred result may still not be close to the actual color transfer result because color transfer methods also adjust contrast and intensity mapping. In our approximation pipeline, the shading adjustment matrix $D$ can be directly obtained from the ALS procedure. When the chromaticity mapping matrix $H$ is estimated from down-sampled images, the estimated $D$ from ALS is not in full-resolution. In this case, according to Equation~\ref{eq:shading}, $D$ can be solved for by a least-squares solution $\min_{D} \norm{DB_{simple}-B}_{F}$. We introduce the additional shading reproduction step as follows.

\subsubsection*{Shading adjustment reproduction -- mapped shading homography}
\begin{figure}[htb!]
    \begin{center}
      \includegraphics[width=\linewidth]{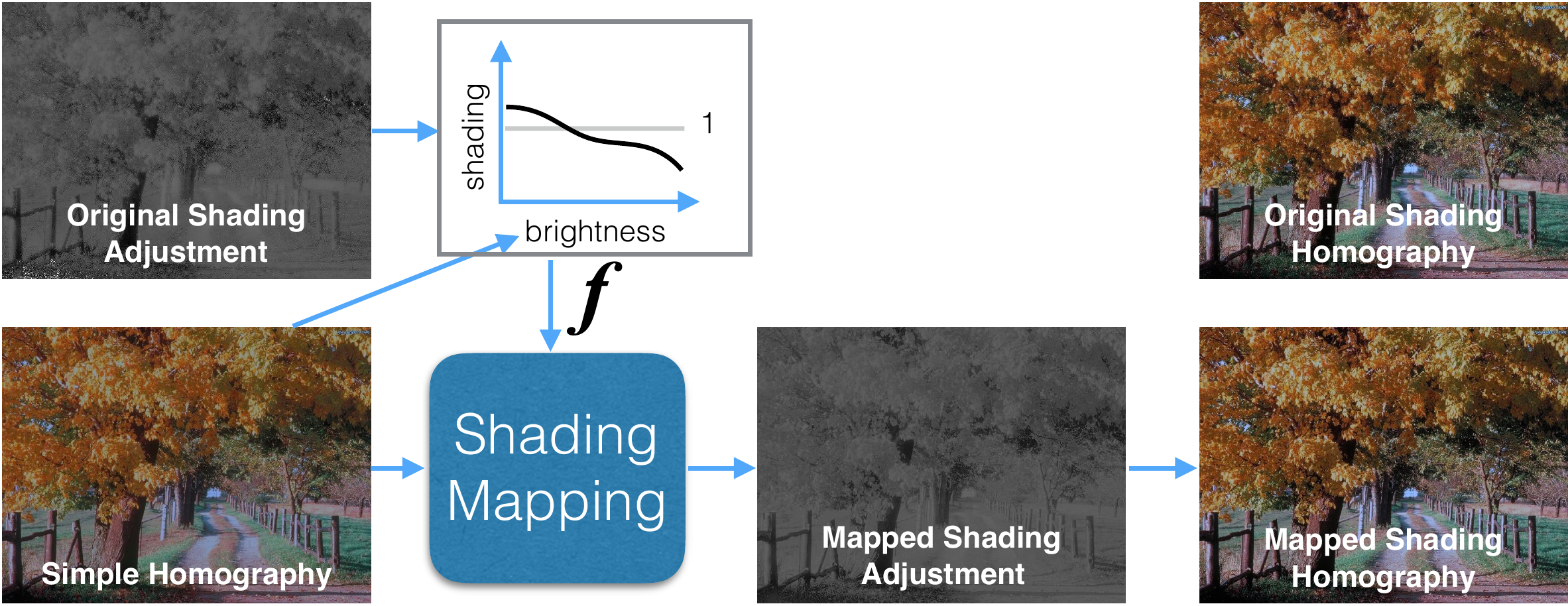}
    \end{center}
    \caption{Shading reproduction. A function $f$ of brightness-to-shading mapping is fitted to the brightness of the simple homography result and the original shading adjustment. The function $f$ is used to reproduce a mapped shading adjustment from which a mapped shading homography result is generated.}
    \label{fig:shading_mapping}
    \vspace{-5pt}
\end{figure}
A universal color transfer decomposition should be compatible with any input image in a similar color theme. Although the chromaticity mapping matrix $H$ is reusable for adjusting other images, the per-pixel shading adjustment $D$ derived from the ALS procedure only works for the source image. To resolve this issue, we propose a brightness-to-shading mapping to reproduce the shading adjustment for any input image. The mapping is estimated by fitting a smooth curve to the per-pixel brightness and shading data. A direct fitting for all data points is computationally costly. Instead, we uniformly divide the brightness range into $50$ slots and compute the center point (average brightness and shading) for each slot. The center point summarizes the point distribution of a range to reduce the amount of data for fitting. The smooth piece-wise curve $f$ is modeled as a Piece-wise Cubic Hermite Interpolating Polynomial (PCHIP)~\cite{pchip} according to the $50$ summarized data sites.

Directly applying the mapped shading adjustment may lead to sharp gradient artifacts because the shading variations for some areas may not follow the global trend. When the shading is not smooth, the overall magnitude of image edges is expected to be large. Inspired by the Laplacian smoothness constraint adopted in \cite{farbman2008edge}, we solve this by minimizing the overall magnitude of Laplacian image edges of the shading image as shown in Equation~\ref{eq:min_lap}:
\begin{equation}
    \min_{D} \norm{D-D_{\text{mapped}}}_{F} + \lambda\norm{I_{D}*K}_{F}
    \label{eq:min_lap}
\end{equation}
where the first term ensures the similarity between the optimum shading $D$ and the $f$-mapped per-pixel shading $D_{\text{mapped}}$, the second term enforces the smoothness constraint, $\lambda$ is a smoothness weight with a small value, $I_{D}$ is the 2D shading image (reshaped from the vector of the diagonal elements of $D$) which is then convolved by a $3\times 3$ Laplacian kernel $K$. See our supplementary material for the details about how to solve this minimization and determine $\lambda$ adaptively.
As is shown in Figure~\ref{fig:shading_mapping}, the shading adjustment can be reproduced according to $f$ and the brightness of the {\it simple} homography transfer result. The result generated from mapped shading is visually close to the original {\it shading} homography result.

\section{Results}
\label{sec:results}
\begin{figure}[htb!]
\begin{center}
  \includegraphics[width=\linewidth]{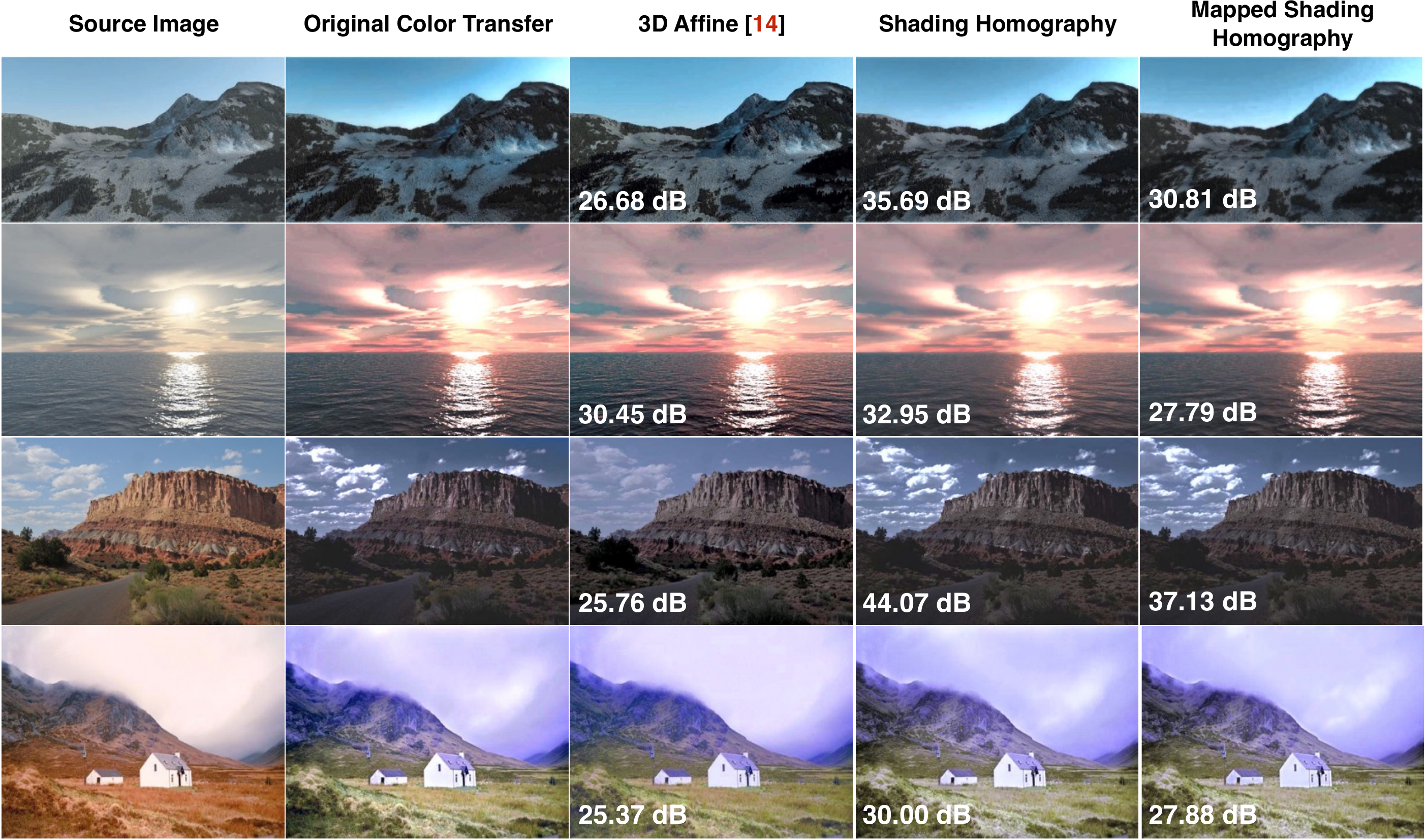}
\end{center}
\caption{Visual result of color transfer approximations (in the order of \cite{Pitie2}, \cite{ReinhardTransfer}, \cite{Pouli}, \cite{Nguyen}). The corresponding PSNR error is shown on each approximation result. Our homography-based methods produce closer approximations to the original color transfer results.}
  \label{fig:visual_comp}
\end{figure}

We first show some visual results of color transfer approximations of \cite{Nguyen,Pitie2,Pouli,ReinhardTransfer} in Figure~\ref{fig:visual_comp}. Global 3D affine mapping~\cite{MKL_ct} does not well reproduce the shading adjustments of color transfer. Our homography-based method offers a closer color transfer approximation. In Table~\ref{tab:rmse}, we also quantitatively evaluate the approximation accuracy of 3 candidates by a PSNR (Peak Signal-to-Noise Ratio) measurement. Acceptable values for wireless image transmission quality loss are considered to be over 20 dB (the higher the better)~\cite{PSNR_im}. The test is based on 7 classic color transfer image pairs and 4 color transfer methods. The original shading homography produces the best result overall. Mapped shading homography also generally produces higher PSNR scores compared with 3D affine mapping, {\it esp.} for \cite{Pitie2,Pouli,ReinhardTransfer}.
\begin{table}[htb!]
\centering
\caption{PSNR measurement between the original color transfer result and its approximation (see our supplementary material for the complete table and their visual results).}
\label{tab:rmse}
\vspace{5pt}
\begin{tabular}{lcccc}
\toprule
 & Nguyen~\cite{Nguyen} & Pitie~\cite{Pitie2} & Pouli~\cite{Pouli} & Reinhard~\cite{ReinhardTransfer} \\ \midrule
3D affine~\cite{MKL_ct} & 26.85 & 26.04 & 26.92 & 28.49 \\
Shading homography& 31.51 & 31.06 & 36.55 & 35.48 \\
Mapped shading homography & 27.77 & 28.16 & 31.70 & 31.18 \\
\bottomrule
\end{tabular}
\end{table}

\section{Applications}
In this section, we show that our color transfer decomposition can be applied to color transfer enhancement and video color grading re-application.
\subsubsection*{Color transfer with reduced artifacts}
\begin{figure}[htb!]
\begin{center}
  \includegraphics[width=\linewidth]{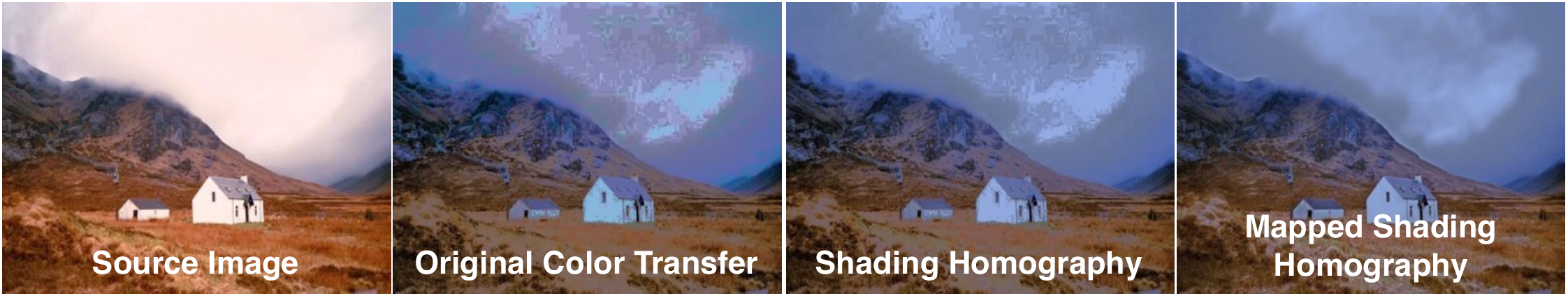}
\end{center}
\caption{Imperfection fixing. The imperfections in a color transfer~\cite{Pouli} and its shading homography approximation are fixed by mapped shading adjustment.}
  \label{fig:fix}
\end{figure}
In Figure~\ref{fig:fix}, original shading homography approximation retains the artifacts of noise and over-saturation of the original color transfer result. These artifacts are fixed by mapped {\it shading} homography.
\subsubsection*{Video color grading re-application}
\begin{figure}[htb!]
\begin{center}
  \includegraphics[width=\linewidth]{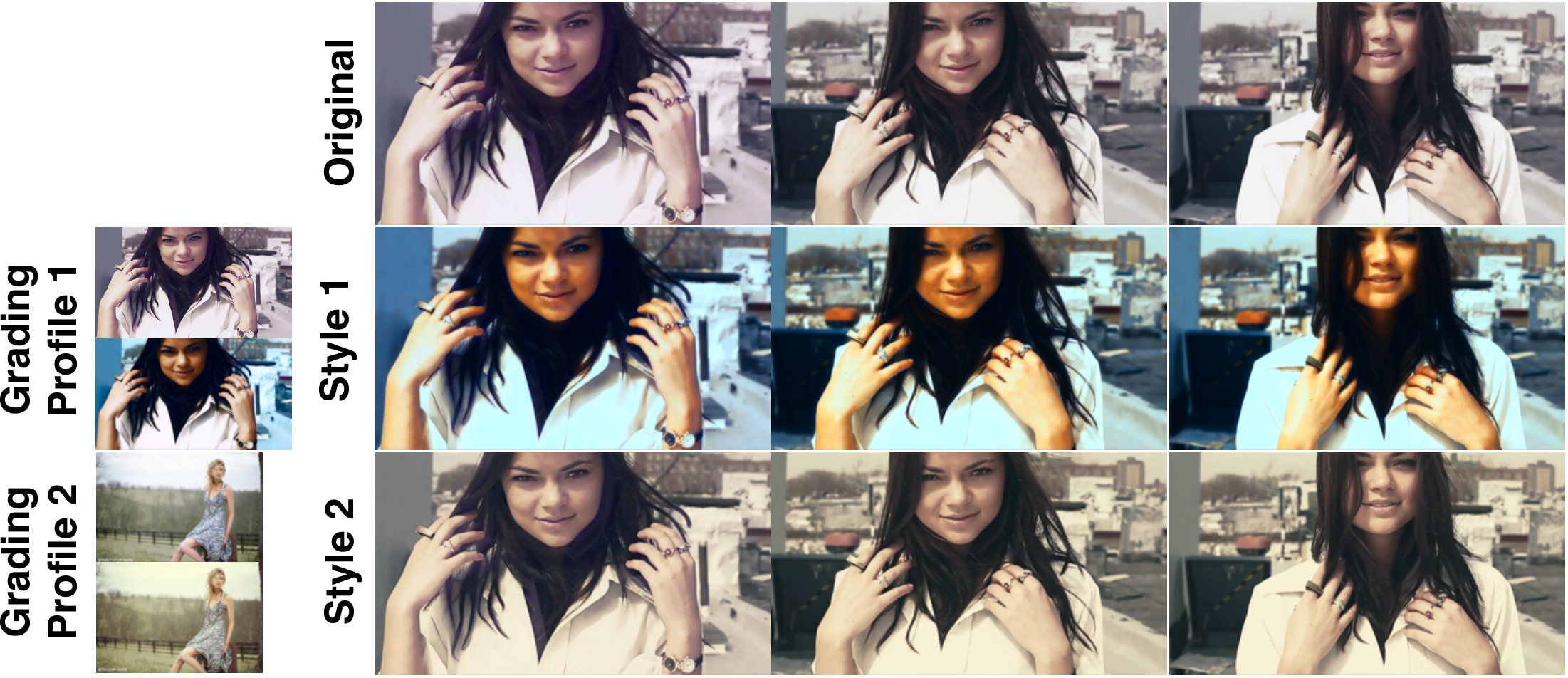}
\end{center}
\caption{Video color grading re-application (original video from Cry \textcopyright Jeffro). The color grading profile is extracted from two pairs of image samples. Grading profile 1 is applied to the same scene. Grading profile 2 is applied to a scene different from its image samples.}
  \label{fig:grading}
\end{figure}
Color transfer methods~\cite{Pitie2,ReinhardTransfer,Pouli,Nguyen} cannot be directly applied to video color grading as per-frame color matching leads to temporal incoherence~\cite{farbman2011tonal,bonneel2013example}. The color homography model is a concise representation of the original complex video color grading adjustments. Compared with a prior art~\cite{bonneel2013example}, our model generates stable results in one-go without the excessive steps for removing artifacts such as flickering and bleeding. As shown in Figure~\ref{fig:grading}, given two sample images profiling the desired color grading adjustment, the complex steps of video color grading can be extracted as a mapped {\it shading} homography transfer. The extracted color grading effect can also be re-applied to a different video sequence of a similar color theme (see our supplementary video for more examples).
\section{Conclusion}
Based on the theorem of color homography, we have shown that a global color transfer can be approximated by a combination of chromaticity mapping and shading adjustment. Our experiment shows that the proposed color transfer decomposition approximates very well many popular color transfer methods. We also demonstrate two applications for fixing the imperfections in a color transfer result and video color grading re-application. We believe that this verified model of color transfer is also important for developing simple and efficient color transfer algorithms.
\label{sec:conclusion}

\section*{Acknowledgment}
This work was supported by EPSRC Grant EP/M001768/1.

%------------------------------------------------------------------------- 

\bibliography{biblio}
\end{document}